\documentclass{article}

\usepackage{times}
\usepackage{tchdr}
\usepackage[accepted]{icml2018}

\algrenewcomment[1]{\hfill $\triangleright$ #1 }
\algnewcommand{\LineComment}[1]{\Statex $\triangleright$ #1 }

\icmltitlerunning{Bayesian Coreset Construction via Greedy Iterative Geodesic Ascent}

\begin{document} 

\twocolumn[
\icmltitle{Bayesian Coreset Construction via Greedy Iterative Geodesic Ascent}

\icmlsetsymbol{equal}{*}

\begin{icmlauthorlist}
\icmlauthor{Trevor Campbell}{csail}
\icmlauthor{Tamara Broderick}{csail}
\end{icmlauthorlist}

\icmlaffiliation{csail}{Computer Science and Artificial Intelligence Laboratory, Massachusetts Institute of Technology, Cambridge, MA, United States}
\icmlcorrespondingauthor{Trevor Campbell}{tdjc@mit.edu}
\icmlkeywords{coreset, Bayesian, inference, scalable, greedy}

\vskip 0.3in
]

\printAffiliationsAndNotice{}  

\begin{abstract}
Coherent uncertainty quantification is a key strength of Bayesian methods.
But modern algorithms for approximate Bayesian posterior inference often sacrifice 
accurate posterior uncertainty estimation in the pursuit of scalability.
This work shows that previous \emph{Bayesian coreset} construction algorithms---which build a small, weighted subset of the data
that approximates the full dataset---are no exception. We demonstrate that 
these algorithms scale the coreset log-likelihood suboptimally, resulting in underestimated posterior uncertainty.
To address this shortcoming, we develop \emph{greedy iterative geodesic ascent (GIGA)}, a novel algorithm for Bayesian coreset construction that scales the coreset log-likelihood optimally. 
GIGA provides geometric decay in posterior approximation error as a function of coreset size, and
maintains the fast running time of its predecessors.
The paper concludes with validation of GIGA on both synthetic and real datasets, demonstrating that it reduces posterior approximation
error by orders of magnitude compared with previous coreset constructions. 
\end{abstract}

\section{Introduction}
Bayesian methods provide
a wealth of options for principled parameter estimation and uncertainty quantification.
But Markov chain Monte Carlo (MCMC) methods \citep{Robert04,Neal11,Hoffman14}, the gold standard for Bayesian inference, typically have complexity $\Theta(NT)$
for dataset size $N$ and number of samples $T$ and are intractable for modern large-scale datasets.
Scalable methods (see \citep{Angelino:2016} for a recent survey), on the other hand, often sacrifice the 
strong guarantees of MCMC and provide unreliable posterior approximations.
For example, variational methods and their scalable and streaming variants
\citep{Jordan99,Wainwright08,Hoffman:2013,Ranganath14,Broderick:2013b,Campbell14,Campbell:2015,Dieng17}
are both susceptible to finding bad local optima in the variational objective and 
tend to either over- or underestimate posterior variance depending on the chosen discrepancy and variational family. 

\begin{figure*}[t!]
\begin{center}
\begin{subfigure}[t]{0.45\textwidth}
\centering\includegraphics[width=\columnwidth, clip, trim=35 30 35 20]{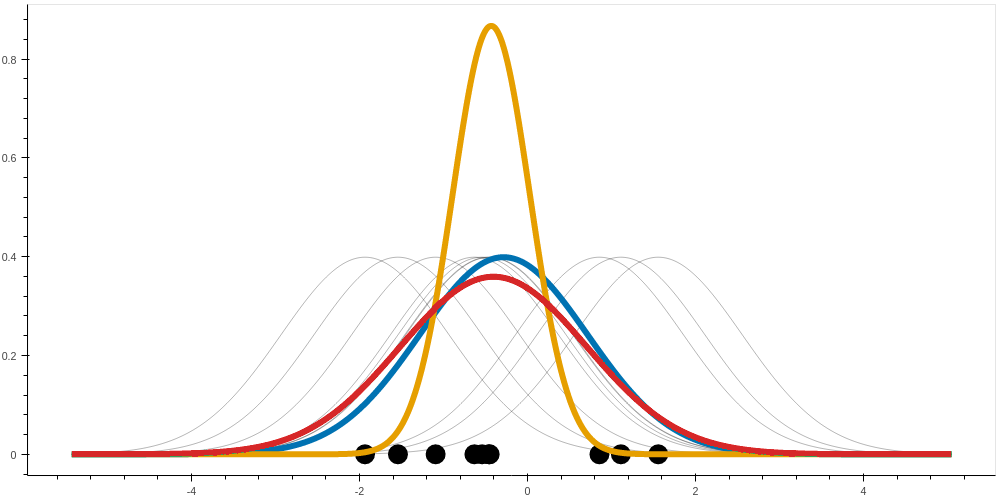}
\end{subfigure}
\begin{subfigure}[t]{0.45\textwidth}
\centering\includegraphics[width=\columnwidth, clip, trim=35 30 35 10]{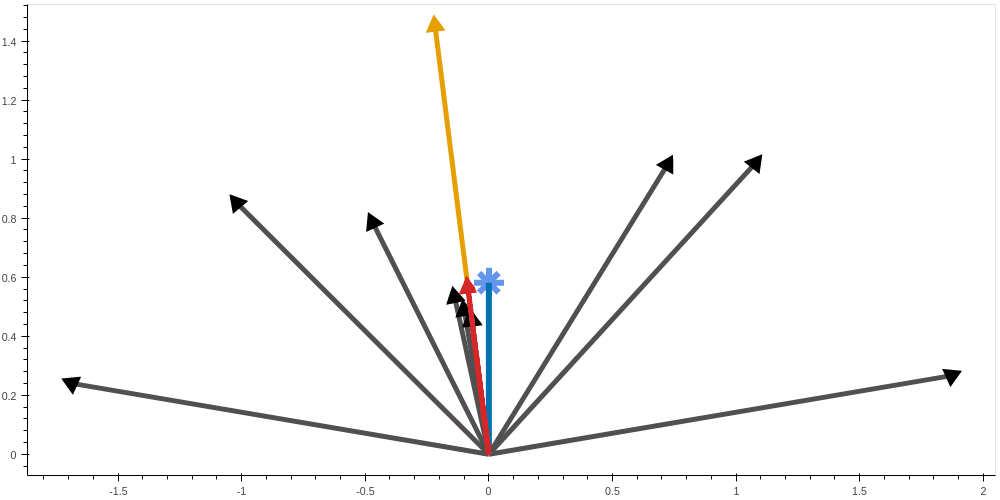}
\end{subfigure}
\caption{
(Left) Gaussian inference for an unknown mean, showing data (black points and likelihood densities),
exact posterior (blue), and optimal coreset posterior approximations of size 1 from solving 
the original coreset construction problem \cref{eq:fullproblem} (red) and the modified problem \cref{eq:modifiedproblem} (orange).
The orange coreset posterior has artificially low uncertainty.
The exact and approximate log-posteriors are scaled down (by the same amount) for visualization.
(Right) The vector formulation of log-likelihoods using the same color scheme.
} \label{fig:problem}
\end{center}
\end{figure*}

Bayesian coresets \citep{Huggins16,Campbell17} provide an alternative 
approach---based on the observation that large datasets often contain redundant data---in which
a small subset of the data of size $M \ll \min\{N, T\}$ is selected and reweighted such that it preserves the statistical properties of the full dataset.
The coreset can be passed to a standard MCMC algorithm, providing
posterior inference with theoretical guarantees at a significantly reduced $O(M(N+T))$ computational cost. 
But despite their advantages, existing Bayesian coreset constructions---like 
many other scalable inference methods---tend to underestimate posterior variance (\cref{fig:problem}).
This effect is particularly evident when the coreset is small, which is the regime 
we are interested in for scalable inference.

In this work, we show that existing Bayesian coreset constructions
underestimate posterior uncertainty because they scale the coreset log-likelihood suboptimally
in order to remain unbiased \citep{Huggins16} or to keep their weights in a particular constraint polytope \citep{Campbell17}.
The result is an overweighted coreset with too much ``artificial data,'' and
therefore an overly certain posterior.
Taking this intuition to its limit, we demonstrate that there exist models for which previous algorithms output coresets with 
arbitrarily large relative posterior approximation error at any coreset size (\cref{prop:fwbad}).
We address this issue by developing a novel coreset construction algorithm, 
\emph{greedy iterative geodesic ascent (GIGA)}, that
optimally scales the coreset log-likelihood to best fit the full dataset log-likelihood.
GIGA has the same computational complexity as the current state of the art, but
its optimal log-likelihood scaling leads to uniformly bounded relative error for \emph{all} models,
as well as asymptotic exponential error decay (\cref{thm:gigabound,cor:gigabound}).
The paper concludes with experimental validation of GIGA on a synthetic vector approximation
problem as well as regression models applied to multiple real and synthetic datasets.

\section{Bayesian Coresets}\label{sec:background}
\subsection{Background}
In Bayesian statistical modeling,
we are given a dataset $\left(y_n\right)_{n=1}^N$ of $N$ observations,
a likelihood $p(y_n | \theta)$ for each observation given a parameter $\theta \in \Theta$,
and a prior density $\pi_0(\theta)$ on $\Theta$.
We assume that the data are conditionally independent given $\theta$.
The Bayesian posterior is given by 
\[
\pi(\theta) &\defined \frac{1}{Z}\exp(\mcL(\theta))\pi_{0}(\theta), \label{eq:bayes}
\]
where the log-likelihood $\mcL(\theta)$ is defined by
\[\<
\mcL_n(\theta) &\defined \log p(y_{n} \given \theta), \quad 
\mcL(\theta) \defined \sum_{n=1}^N\mcL_n(\theta),\label{eq:bayesdefns}
\>\]
and $Z$ is the marginal likelihood. 
MCMC returns approximate samples from the posterior, which can be used to construct
an empirical approximation to the posterior distribution. Since each sample
requires at least one full likelihood evaluation---typically an $\Theta(N)$ operation---MCMC has $\Theta(NT)$ complexity for $T$ posterior samples.

To reduce the complexity, we can instead run MCMC on a \emph{Bayesian coreset} \citep{Huggins16}, a small, weighted subset of the data. 
Let $w \in \reals^N$ be the vector of nonnegative weights, with weight $w_n$ for data point $y_n$,
and $\|w\|_0 \defined \sum_{n=1}^N\ind\left[w_n>0\right] \ll N$.
Then we approximate the full log-likelihood with
$\mcL(w, \theta) \defined \sum_{n=1}^N w_n \mcL_n(\theta)$ and run MCMC with the approximated likelihood.
By viewing the log-likelihood functions $\mcL_n(\theta), \mcL(\theta), \mcL(w, \theta)$ as vectors $\mcL_n, \mcL, \mcL(w)$ in a normed vector space,
\citet{Campbell17} pose the problem of constructing a coreset of size $M$ as
cardinality-constrained vector approximation,
\[\<
\min_{w \in \reals^N} \quad & \left\|\mcL(w) - \mcL \right\|^2 \\
 \text{s.t.} \quad  & w \geq 0, \quad \|w\|_0 \leq M.\label{eq:fullproblem}
\>\]
Solving \cref{eq:fullproblem} exactly is not tractable for large $N$ due to the cardinality
constraint; approximation is required. 
Given a norm induced by an inner product, and defining 
\[
\sigma_n \defined \left\|\mcL_n\right\| \quad\text{and}\quad \sigma \defined \sum_{n=1}^N \sigma_n,
\]
\citet{Campbell17} replace the cardinality constraint in \cref{eq:fullproblem} 
with a simplex constraint,
\[\<
\min_{w \in \reals^N} \quad & \left\|\mcL(w) - \mcL\right\|^2 \\
\text{s.t.} \quad & w\geq0, \quad \sum_{n=1}^N \sigma_n w_n = \sigma.\label{eq:modifiedproblem}
\>\]
\cref{eq:modifiedproblem} can be solved while ensuring $\|w\|_0 \leq M$ 
using either importance sampling (IS) or Frank--Wolfe (FW) \citep{Frank56}. 
Both procedures add one data point to the linear combination $\mcL(w)$ 
at each iteration; IS chooses the new data point \iid with probability $\nicefrac{\sigma_n}{\sigma}$,
while FW chooses the point most aligned with the residual error.
This difference in how the coreset is built results in different convergence behavior:
FW exhibits geometric convergence $\left\|\mcL(w) - \mcL\right\|^2 = O(\nu^M)$ for some $0<\nu < 1$,
while IS is limited by the Monte Carlo rate $\left\|\mcL(w) - \mcL\right\|^2 = O(M^{-1})$ with high probability \citep[Theorems 4.1, 4.4]{Campbell17}.
For this reason, FW is the preferred method for coreset construction.

\cref{eq:fullproblem} is a special case of the sparse vector approximation problem, 
which has been studied extensively in past literature. 
Convex optimization formulations---e.g.~basis pursuit \citep{Chen99},
LASSO \citep{Tibshirani96}, 
the Dantzig selector \citep{Candes07},
and compressed sensing \citep{Candes05,Donoho06,Boche15}---are 
expensive to solve compared to our greedy approach, 
and often require tuning regularization coefficients and thresholding to ensure cardinality constraint feasibility.
Previous greedy iterative algorithms---e.g.~(orthogonal) matching pursuit \citep{Mallat93,Chen89,Tropp04},
Frank--Wolfe and its variants \citep{Frank56,Guelat86,Jaggi13,LacosteJulien15,Locatello17}, 
Hilbert space vector approximation methods \citep{Barron08}, 
kernel herding \citep{Chen10},
and AdaBoost \citep{Freund97}---have sublinear 
error convergence unless computationally expensive correction steps are included.
In contrast, the algorithm developed in this paper has 
no correction steps, no tuning parameters, and geometric error convergence. 

\subsection{Posterior Uncertainty Underestimation}

The new $\sum_n \sigma_nw_n = \sigma$ constraint in \cref{eq:modifiedproblem} 
has an unfortunate practical consequence: both IS and FW must scale the coreset log-likelihood suboptimally---roughly, by $\sigma$ rather than $\|\mcL\|$ as 
they should---in order to maintain feasibility. Since $\sigma \geq \|\mcL\|$, intuitively the coreset construction algorithms
are adding too much ``artificial data'' via the coreset weights, resulting in an overly certain posterior approximation. 
It is worth noting that this effect is apparent in the error bounds developed by \citet{Campbell17},
which are all proportional to $\sigma$ rather than $\|\mcL\|$ as one might hope for when approximating $\mcL$.

\cref{fig:problem} provides intuition in the setting of Gaussian inference 
for an unknown mean. In this example, we construct the optimal coreset of size 1 for the modified problem in \cref{eq:modifiedproblem} (orange)
and for the original coreset construction problem that we would ideally like to solve in \cref{eq:fullproblem} (red). Compared with the exact posterior (blue),
the orange coreset approximation from \cref{eq:modifiedproblem} has artificially low uncertainty, since it must place weight $\nicefrac{\sigma}{\sigma_n}$ on 
its chosen point. 
Building on this intuition, \cref{prop:fwbad} shows that there are
problems\footnote{While the proof uses orthogonal vectors in $\reals^N$ for simplicity, a similar 
construction arises from the model  $\theta\dist\distNorm(0, I)$,  $x_n\dist\distNorm(\theta_n, 1)$, $n\in[N]$
given the norm $\|\mcL\| = \sqrt{\EE_\pi\left[\|\grad\mcL\|^2\right]}$.} 
for which both FW and IS perform arbitrarily poorly
for any number of iterations $M$. 
\bnprop\label{prop:fwbad}
For any $M\in\nats$, there exists $\left(\mcL_n\right)_{n=1}^N$ for which both
the FW and IS coresets after $M$ iterations have arbitrarily large error relative to $\|\mcL\|$.
\enprop
\bprf
Let $\mcL_n = \nicefrac{1}{N}1_n$ where $1_n$ is the indicator
for the $n^\text{th}$ component of $\reals^N$,
and let $\mcL = \sum_{n=1}^N \mcL_n$. Then in the 2-norm, $\sigma_n = \nicefrac{1}{N}$,
$\sigma = 1$, and  $\left\|\mcL\right\| = \nicefrac{1}{\sqrt{N}}$.
By symmetry, the optimal $w$
for \cref{eq:modifiedproblem} satisfying $\|w\|_0 \leq M$
has uniform nonzero weights $\nicefrac{N}{M}$. Substituting yields
\[
\frac{\left\|\mcL(w) - \mcL\right\|}{\left\|\mcL\right\|}
 &= \sqrt{\frac{N}{M}-1},
\]
which can be made as large as desired by increasing $N$.
The result follows since both FW and IS generate a feasible solution $w$ for \cref{eq:modifiedproblem} satisfying $\|w\|_0 \leq M$.
\eprf
In contrast, the red coreset approximation obtained by solving \cref{eq:fullproblem} scales its weight to minimize $\|\mcL(w) - \mcL\|$, resulting in 
a significantly better approximation of posterior uncertainty. Note that we can scale the weight vector $w$ by 
any $\alpha \geq 0$ without affecting feasibility in the cardinality constraint, i.e., $\|\alpha w\|_0 \leq \|w\|_0$.
In the following section, we use this property to develop a greedy coreset construction algorithm that, unlike FW and IS, maintains optimal 
log-likelihood scaling in each iteration.

\section{Greedy Iterative Geodesic Ascent (GIGA)}\label{sec:hilbert}

In this section, we provide a new algorithm for Bayesian coreset construction and demonstrate that it yields improved approximation error guarantees proportional to $\|\mcL\|$ 
(rather than $\sigma \geq \|\mcL\|$). We begin in \cref{sec:radialopt} by solving
for the optimal log-likelihood scaling analytically. After solving this ``radial optimization problem,''
we are left with a new optimization problem on the unit hyperspherical manifold. 
In \cref{sec:initialization,sec:iteration}, we demonstrate how to solve this new problem by
iteratively building the coreset one point at a time.
Since this procedure selects the point greedily based on a geodesic alignment criterion,
%
%
we call it \emph{greedy iterative geodesic ascent (GIGA)}, detailed in \cref{alg:giga}.
In \cref{sec:output} we scale the resulting coreset optimally using the procedure developed in \cref{sec:radialopt}.
Finally, in \cref{sec:convergence}, we prove that \cref{alg:giga} provides approximation error that is
proportional to $\|\mcL\|$ and geometrically decaying in $M$, as shown by \cref{thm:gigabound}.
\bnthm \label{thm:gigabound}
The output $w$ of \cref{alg:giga} satisfies
\[
\left\|\mcL(w) - \mcL\right\| &\leq \eta \|\mcL\|\nu_M,
\]
where $\nu_M$ is decreasing and $\leq 1$ for all $M\in\nats$, $\nu_M = O(\nu^M)$ for some $0 < \nu < 1$, and $\eta$ is defined by
\[
0 \leq \eta &\defined \sqrt{1-\left(\max_{n\in[N]}\left< \frac{\mcL_n}{\|\mcL_n\|}, \frac{\mcL}{\|\mcL\|}\right>\right)^2} \leq 1.\label{eq:etadefn}
\]
\enthm
The particulars of the sequence $\left(\nu_M\right)_{M=1}^\infty$ are somewhat involved; see \cref{sec:convergence} for the detailed development.
A straightforward consequence of \cref{thm:gigabound} is that the issue described in \cref{prop:fwbad}
has been resolved; the solution $\mcL(w)$ is always scaled optimally, leading to decaying relative error.
\cref{cor:gigabound} makes this notion precise.
\bncor\label{cor:gigabound}
For any set of vectors $\left(\mcL_n\right)_{n=1}^N$, \cref{alg:giga}
provides a solution to \cref{eq:fullproblem} with error $\leq 1$ relative to $\|\mcL\|$.
\encor

\begin{algorithm}[t!]
\caption{GIGA: Greedy Iterative Geodesic Ascent}\label{alg:giga}
\begin{algorithmic}[1]
\Require $(\mcL_n)_{n=1}^N$, $M$, $\left<\cdot, \cdot\right>$ 
\LineComment{Normalize vectors and initialize weights to 0 }
\State $\mcL \gets \sum_{n=1}^N \mcL_n$ 
\State $\forall n \in [N] \, \, \ell_n \gets \frac{\mcL_n}{\|\mcL_n\|}$,  $\ell \gets \frac{\mcL}{\|\mcL\|}$
\State $w_0 \gets 0$, $\ell(w_0) \gets 0$
\For{$t \in \{0, \dots, M-1\}$}
\LineComment{Compute the geodesic direction for each data point }
\State $d_t \gets \frac{\ell - \left<\ell, \ell(w_t)\right>\ell(w_t)}{\|\ell - \left<\ell, \ell(w_t)\right>\ell(w_t)\|}$
\State  $\forall n \in [N]$, $d_{tn} \gets \frac{\ell_n - \left<\ell_n, \ell(w_t)\right>\ell(w_t)}{\|\ell_n - \left<\ell_n, \ell(w_t)\right>\ell(w_t)\|}$
\LineComment{Choose the best geodesic }
\State $n_t \gets \argmax_{n \in [N]} \left<d_t, d_{tn}\right>$
\State $\zeta_0 \gets \left<\ell, \ell_{n_t}\right>$, $\zeta_1 \gets \left<\ell, \ell(w_t)\right>$, $\zeta_2 \gets\left<\ell_{n_t}, \ell(w_t)\right>$
\LineComment{Compute the step size }
\State $\gamma_t \gets \frac{\zeta_0 - \zeta_1\zeta_2}{(\zeta_0 - \zeta_1\zeta_2) + (\zeta_1 - \zeta_0\zeta_2)}$
\LineComment{Update the coreset }
\State $\ell(w_{t+1}) \gets \frac{(1-\gamma_t)\ell(w_t) + \gamma_t\ell_{n_t}}{\|(1-\gamma_t)\ell(w_t) + \gamma_t\ell_{n_t}\|}$
\State $w_{t+1} \gets \frac{(1-\gamma_t) w_t + \gamma_t1_{n_t}}{\|(1-\gamma_t)\ell(w_t) + \gamma_t\ell_{n_t}\|}$
\EndFor
\LineComment{Scale the weights optimally }
\State $\forall n \in [N]$, $w_{Mn} \gets w_{Mn} \frac{\|\mcL\|}{\|\mcL_n\|}\left<\ell(w_M), \ell\right>$
\State \Return $w$
\end{algorithmic}
\end{algorithm}

\subsection{Solving the Radial Optimization}\label{sec:radialopt}
We begin again with the problem of coreset construction for a collection of vectors $\left(\mcL_n\right)_{n=1}^N$
given by \cref{eq:fullproblem}. Without loss of generality, we assume $\|\mcL\| > 0$ and $\forall n \in [N]$, $\|\mcL_n\| > 0$;
if $\|\mcL\| = 0$ then $w=0$ is a trivial optimal solution, and any $\mcL_n$ for which $\|\mcL_n\| = 0$ can 
be removed from the coreset without affecting the objective in \cref{eq:fullproblem}.
Recalling that the weights of a coreset can be scaled by an arbitrary constant $\alpha\geq 0$ without affecting feasibility,
we rewrite \cref{eq:fullproblem} as
\[
\<
\min_{\alpha \in \reals, w\in\reals^N} \quad & \left\|\alpha \mcL(w) - \mcL\right\|^2\\
\mathrm{s.t.} \quad & \alpha \geq 0, \, w \geq 0, \, \|w\|_0 \leq M.
\>\label{eq:radialproblem}
\]
Taking advantage of the fact that $\|\cdot\|$ is induced by an inner product, we
can expand the objective in \cref{eq:radialproblem} as a quadratic function of $\alpha$,
and analytically solve the optimization in $\alpha$ to yield
\[
\alpha^\star = \frac{\|\mcL\|}{\|\mcL(w)\|}\max\left\{0, \left<\frac{\mcL(w)}{\|\mcL(w)\|}, \frac{\mcL}{\|\mcL\|}\right>\right\}.\label{eq:optimalscaling}
\]
In other words, $\mcL(w)$ should be rescaled to have norm $\|\mcL\|$, and then scaled further depending on its directional
alignment with $\mcL$.  We substitute this result back into \cref{eq:radialproblem} to find that coreset construction is equivalent
to solving 
\[
\<
\hspace{-.2cm}\min_{w\in\reals^N} \quad & \|\mcL\|^2\left( 1 - \max\left\{0, \left<\frac{\mcL(w)}{\|\mcL(w)\|}, \frac{\mcL}{\|\mcL\|}\right>\right\}^2\right)\\
\mathrm{s.t.} \quad & w \geq 0, \, \|w\|_0 \leq M.
\>\label{eq:scaledproblem}
\]
Note that by selecting the optimal scaling via \cref{eq:optimalscaling}, 
the cost now depends only on the alignment of $\mcL(w)$ and $\mcL$ irrespective of their norms.
Finally, defining the normalized vectors $\ell_n \defined \frac{\mcL_n}{\|\mcL_n\|}$, $\ell \defined \frac{\mcL}{\|\mcL\|}$, and 
$\ell(w) \defined \sum_{n=1}^N w_n \ell_n$, solving \cref{eq:scaledproblem} is equivalent
to maximizing the alignment of $\ell$ and $\ell(w)$ over $\ell(w)$ lying on the unit hypersphere: 
\[
\<
\max_{w\in\reals^N} \quad & \left<\ell(w), \ell\right>\\
\mathrm{s.t.} \quad & w \geq 0, \, \|w\|_0 \leq M, \, \|\ell(w)\| = 1.
\>\label{eq:geodesiccoresetproblem}
\]
Compare this optimization problem to \cref{eq:modifiedproblem}; \cref{eq:geodesiccoresetproblem}
shows that the natural choice of manifold for optimization is the \emph{unit hypersphere}
rather than the simplex. Therefore, in \cref{sec:initialization,sec:iteration}, we develop a 
greedy optimization algorithm that operates on the hyperspherical manifold.
At each iteration, the algorithm picks the point indexed by $n$ for which the geodesic between $\ell(w)$ and $\ell_n$
is most aligned with the geodesic between $\ell(w)$ and $\ell$. Once the point has been added, it reweights the coreset
and iterates.

\subsection{Coreset Initialization}\label{sec:initialization}
For the remainder of this section, denote the value of the weights at iteration $t$ by $w_t \in \reals^N_+$,
and write the $n^\text{th}$ component of this vector as $w_{tn}$.
To initialize the optimization, we select the vector $\ell_n$ most aligned with $\ell$, i.e.
\[
n_1  = \argmax_{n\in[N]} \left<\ell_n, \ell\right>,  \quad w_1 = 1_{n_1}.
\]
This initialization provides two benefits: for all iterations $t$, there is no point $\ell_n$
in the hyperspherical cap $\{y : \|y\|=1, \, \left<y, \ell\right> \geq r\}$ 
for any $r > \left<\ell(w_1), \ell\right>$, as well as $\left<\ell(w_1), \ell\right> > 0$ by \cref{lem:init}.
We use these properties to develop the error bound in \cref{thm:gigabound} and guarantee that the iteration step developed below in \cref{sec:iteration} is 
always feasible.  
Note that \cref{alg:giga} does not explicitly run this initialization and simply sets $w_0=0$, since
the initialization for $w_1$ is equivalent to the usual iteration step described in \cref{sec:iteration} after setting $w_0=0$.
\bnlem\label{lem:init}
The initialization satisfies 
\[
\left<\ell(w_1), \ell\right> \geq \frac{\|\mcL\|}{\sigma} > 0.
\]
\enlem
\bprf
For any $\xi \in \Delta^{N-1}$, we have that
\[
\left<\ell(w_1), \ell\right> = \max_{n\in[N]} \left<\ell_n, \ell\right> \geq \sum_n \xi_n \left<\ell_n, \ell\right>.
\]
So choosing $\xi_n = \sigma_n/\sigma$,
\[
\hspace{-.3cm}\left<\ell(w_1), \ell\right> \geq \frac{1}{\sigma} \left<\sum_n \mcL_n, \ell\right> = \frac{\|\mcL\|}{\sigma} \left<\ell, \ell\right> = \frac{\|\mcL\|}{\sigma}.
\]
By assumption (see \cref{sec:radialopt}), we have $\|\mcL\| > 0$.
\eprf

\subsection{Greedy Point Selection and Reweighting}\label{sec:iteration}
\begin{figure}[t!]
\begin{center}
\begin{subfigure}[t]{0.47\columnwidth}
\centering\includegraphics[width=.9\columnwidth, clip, trim=30 30 30 30]{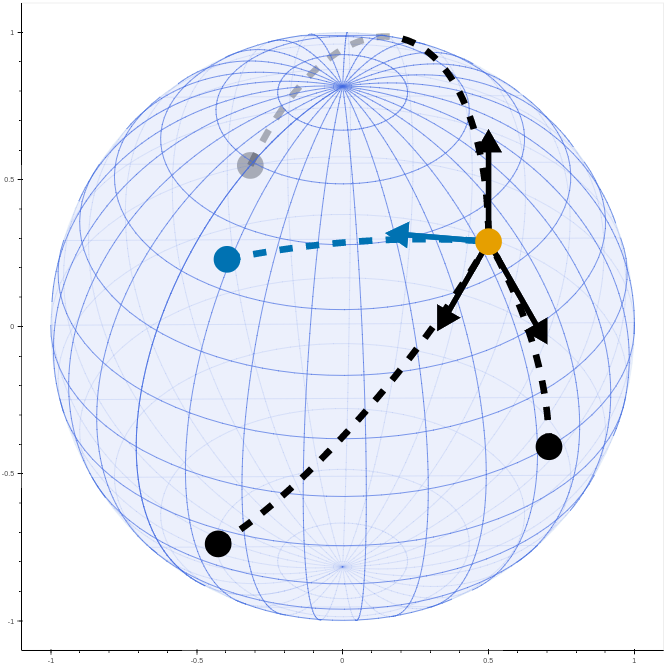}
\caption{}\label{fig:greedya}
\end{subfigure}
\begin{subfigure}[t]{0.47\columnwidth}
\centering\includegraphics[width=.9\columnwidth, clip, trim=30 30 30 30]{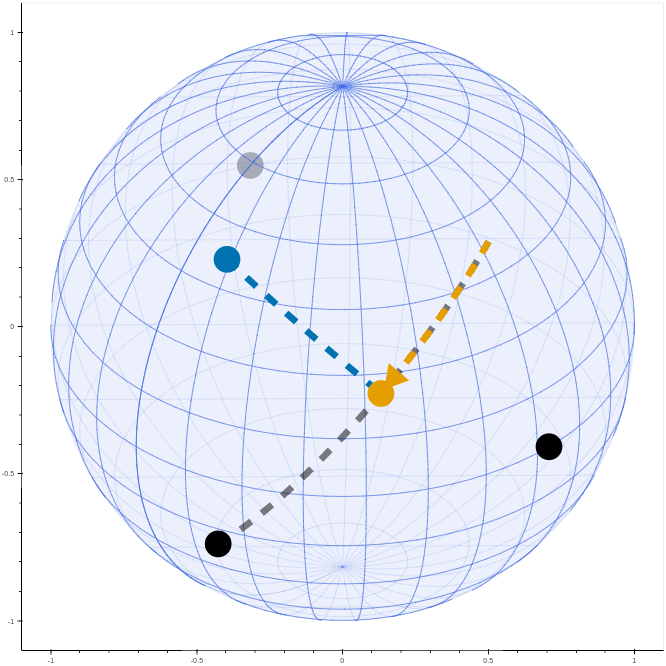}
\caption{}\label{fig:greedyb}
\end{subfigure}
\end{center}
\caption{\cref{fig:greedya} shows the greedy geodesic selection procedure, with normalized total log-likelihood $\ell$ in blue, 
normalized data log-likelihoods $\ell_n$ in black, and the current iterate $\ell(w)$ in orange. 
\cref{fig:greedyb} shows the subsequent line search procedure.}
\label{fig:greedy}
\end{figure}

At each iteration $t$, we have a set of weights $w_t\in\reals_+^N$ for which $\|\ell(w_t)\| = 1$
and $\|w_t\|_0 \leq t$. The next point to add to the coreset is the one for which the geodesic direction
is most aligned with the direction of the geodesic from $\ell(w_t)$ to $\ell$; 
\cref{fig:greedya} illustrates this selection.
Precisely, we define
\[
\<
d_t &\defined \frac{\ell-\left<\ell, \ell(w_t)\right>\ell(w_t)}{\|\ell-\left<\ell, \ell(w_t)\right>\ell(w_t)\|}\\
\forall n\in[N], \quad d_{tn}&\defined \frac{\ell_n-\left<\ell_n, \ell(w_t)\right>\ell(w_t)}{\|\ell_n-\left<\ell_n, \ell(w_t)\right>\ell(w_t)\|},
\>\label{eq:tangentvecs}
\]
and we select the data point at index $n_t$ where
\[
n_t &= \argmax_{n\in[N]} \left<d_t, d_{tn}\right>.\label{eq:selection}
\]
This selection leads to the largest decrease in approximation error, as shown later in \cref{eq:costrecursion}.
For any vector $u$ with $\|u\| = 0$, we define $\frac{u}{\|u\|} \defined 0$ to avoid any issues
with \cref{eq:selection}.
Once the vector $\ell_{n_t}$ has been selected, the next task is to redistribute the weights
between $\ell(w_t)$ and $\ell_{n_t}$ to compute $w_{t+1}$. We perform a search along the geodesic
from $\ell(w_t)$ to $\ell_{n_t}$ to maximize $\left<\ell(w_{t+1}), \ell\right>$, as shown in \cref{fig:greedyb}:
\[
\gamma_t &= \argmax_{\gamma\in[0,1]} \left<\ell, \frac{\ell(w_t)+\gamma(\ell_{n_t}-\ell(w_t))}{\|\ell(w_t)+\gamma(\ell_{n_t}-\ell(w_t))\|}\right>.\label{eq:linesearch}
\]
Constraining $\gamma \in [0, 1]$ ensures that the resulting $w_{t+1}$ is feasible for \cref{eq:geodesiccoresetproblem}. 
Taking the derivative and setting it to 0 yields the unconstrained optimum of \cref{eq:linesearch},
\[
\hspace{-.2cm}\zeta_0 &\defined \left<\ell, \ell_{n_t}\right> \,\, \zeta_1 \defined \left<\ell, \ell(w_t)\right> \,\, \zeta_2 \defined\left<\ell_{n_t}, \ell(w_t)\right>\\
\gamma_t &\phantom{:}= \frac{\left(\zeta_0 - \zeta_1\zeta_2\right)}{\left(\zeta_0-\zeta_1\zeta_2\right) + \left(\zeta_1 - \zeta_0\zeta_2\right)}.
\]

\cref{lem:linesearch} shows that $\gamma_t$ is also the solution to the constrained line search problem.
The proof of \cref{lem:linesearch} is based on the fact that 
 we initialized the procedure with the vector most aligned with $\ell$ (and so $\ell_{n_t}$ must lie on the ``other side'' of $\ell$ from $\ell(w_t)$, i.e.~$\gamma_t \leq 1$),
along with the fact that the optimal objective in \cref{eq:selection} is guaranteed to be positive (so moving towards $\ell_{n_t}$ is guaranteed to improve the objective, i.e.~$\gamma_t \geq 0$).
\bnlem\label{lem:linesearch}
For all $t\in\nats$, $\gamma_t \in [0, 1]$.
\enlem
\bprf
It is sufficient to show that both $\zeta_0 - \zeta_1\zeta_2$ and $\zeta_1 - \zeta_0\zeta_2$ are nonnegative and that at least one is strictly positive.
First, examining \cref{eq:selection}, note that for any $\xi\in\Delta^{N-1}$,
\[
\left<d_t, d_{tn_t}\right> = \max_{n\in[N]}\left<d_t, d_{tn}\right> &\geq \sum_{n=1}^N\xi_n\left<d_t, d_{tn}\right>,
\]
so choosing $\xi_n = C \|\mcL_n\|\|\ell_n-\left<\ell_n, \ell(w_t)\right>\ell(w_t)\|$ where $C>0$ is chosen appropriately to normalize $\xi$, we have that
\[
 \left<d_t, d_{tn_t}\right> &\geq C\|\mcL\|\|\ell - \left<\ell, \ell(w_t)\right>\ell(w_t)\| > 0,
\]
since $\|\mcL\|>0$ by assumption (see \cref{sec:radialopt}), and $\ell \neq \ell(w_t)$ since otherwise 
we could terminate the optimization after iteration $t-1$. 
By expanding the formulae for $d_t$ and $d_{tn_t}$, we have that
for some $C'>0$,
\[
\zeta_0 - \zeta_1\zeta_2 = C'\left<d_t, d_{tn_t}\right> > 0.\label{eq:zprop1}
\]
For the other term, first note that $\left<\ell, \ell(w_t)\right>$ is a monotonically increasing function in $t$ 
since $\gamma = 0$ is feasible in \cref{eq:linesearch}. Combined with the choice
of initialization for $w_1$, we have that $\zeta_1 \geq \zeta_0$.  
Further, since $\zeta_2 \in [-1, 1]$ and $\zeta_1 \geq 0$ by its initialization,
\cref{eq:zprop1} implies $\zeta_1 \geq \zeta_1(-\zeta_2) \geq -\zeta_0$.
Therefore $\zeta_1 \geq |\zeta_0| \geq |\zeta_0\zeta_2| \geq \zeta_0\zeta_2$, and the proof is complete.
%
\eprf

Given the line search step size $\gamma_t$, we can now update the weights to compute $\ell(w_{t+1})$ and $w_{t+1}$:
\[
\ell(w_{t+1}) = \frac{\ell(w_t)+\gamma_t(\ell_{n_t}-\ell(w_t))}{\|\ell(w_t)+\gamma_t(\ell_{n_t}-\ell(w_t))\|}\\
w_{t+1} = \frac{w_t + \gamma_t(1_{n_t} - w_t)}{\|\ell(w_t)+\gamma_t(\ell_{n_t}-\ell(w_t))\|}.
\]
There are two options for practical implementation: either (A) we keep track of only $w_t$, recomputing and normalizing $\ell(w_t)$ at each
iteration to obtain $w_{t+1}$; or (B) we keep track of and store both $w_t$ and $\ell(w_t)$. 
In theory both are equivalent, but in practice
(B) will cause numerical errors to accumulate, making $\ell(w_t)$ not equal to the sum of normalized $\ell_n$ weighted by $w_{tn}$.
However, in terms of vector sum and inner-product operations, (A) incurs an additional cost of 
$O(t)$ at each iteration $t$---resulting in a total of $O(M^2)$ for $M$ iterations---while (B) has constant cost for each iteration
with a total cost of $O(M)$. We found empirically that option (B) is preferable in practice, since the reduced numerical error of (A) does not justify the $O(M^2)$ cost.

\subsection{Output}\label{sec:output}
After running $M$ iterations, the algorithm must output a set of weights that correspond to unnormalized vectors $\mcL_n$ rather
than the normalized counterparts $\ell_n$ used within the algorithm. In addition, weights need to be rescaled optimally by 
$\alpha^\star$ from \cref{eq:optimalscaling}. The following formula adjusts the weights $w_M$ 
accordingly to generate the output:
\[
\forall n \in [N], \quad w_{Mn} \gets w_{Mn}\frac{\|\mcL\|}{\|\mcL_n\|}\left<\ell(w_M), \ell\right>.\label{eq:output}
\]

\subsection{Convergence Guarantee}\label{sec:convergence}
We now prove \cref{thm:gigabound} by bounding the objective
of \cref{eq:scaledproblem} as a function of the number of iterations $M$ of greedy geodesic ascent.
Writing $J_t \defined 1-\left<\ell(w_t), \ell\right>^2$ for brevity, we have that 
$\|\mcL\|^2J_t = \|\alpha^\star \mcL(w_t) - \mcL\|^2$ for $\alpha^\star$ from \cref{eq:optimalscaling},
so providing an upper bound on $J_t$ provides a relative bound on the coreset error. Substituting
 the optimal line search value $\gamma_t$ into the objective of \cref{eq:linesearch} yields
\[
J_{t+1} &= J_t\left(1 - \left<d_t, d_{tn_t}\right>^2\right),\label{eq:costrecursion}
\]
where $d_t$, $d_{tn}$, and $n_t$ are given in \cref{eq:tangentvecs,eq:selection}.
In other words, the cost $J_t$ decreases corresponding to how well aligned 
the $\ell(w_t)$-to-$\ell$ and $\ell(w_t)$-to-$\ell_{n_t}$ geodesics are.
We define for each $n\in[N]$ the geodesic direction from $\ell$ to $\ell_n$ to be $d_{\infty n} \defined \frac{\ell_n - \left<\ell_n, \ell\right>\ell}{\|\ell_n - \left<\ell_n, \ell\right>\ell\|}$; this slight abuse of notation from \cref{eq:tangentvecs} is justified since we expect $\ell(w_t) \to \ell$ as $t\to\infty$.
Then the worst-case alignment is governed by two constants, $\tau$ and $\epsilon$:
\[
\tau^{-1} &\defined \min_{s\in\reals^N} \left\|s\right\|_1 \quad \text{s.t.} \quad \ell = \sum_{m=1}^N s_m\ell_m \quad s \geq 0\label{eq:taudefn}\\
\epsilon &\defined \min_{s\in\reals^N} \max_{n\in[N]} \left<-\sum_{m}s_m d_{\infty m}, d_{\infty n}\right>\label{eq:epsdefn}\\
&\hspace{.75cm}\text{s.t.}\quad \|\sum_{m}s_m d_{\infty m}\| = 1, \quad s \geq 0.\notag
\]
Both are guaranteed to be positive by \cref{lem:constantbounds} below.
The constant $\tau$ captures how fundamentally hard it is to approximate $\ell$ using $(\ell_n)_{n=1}^N$, and governs the worst-case behavior
of the large steps taken in early iterations. 
In contrast, $\epsilon$ captures the worst-case behavior of the smaller steps in the
$t\to\infty$ asymptotic regime when $\ell(w_t) \approx \ell$.
These notions are quantified precisely in \cref{lem:geodesiclowerbound}.
The proofs of both \cref{lem:constantbounds,lem:geodesiclowerbound} may be found in
\cref{sec:proofs}.
\bnlem\label{lem:constantbounds}
The constants $\tau$ and $\epsilon$ satisfy
\[
\tau \geq \frac{\|\mcL\|}{\sigma} > 0 &&\text{and}&& \epsilon > 0.
\]
\enlem
\bnlem\label{lem:geodesiclowerbound}
The geodesic alignment $\left<d_t, d_{tn_t}\right>$ satisfies
\[
\left<d_t, d_{tn_t}\right> \geq \tau\sqrt{J_t} \vee f(J_t),
\]
where $f: [0, 1] \to \reals$ is defined by
\[
f(x) &\defined 
\frac{\sqrt{1-x}\sqrt{1-\beta^2} \epsilon + \sqrt{x}\beta}{\sqrt{1-\left(\sqrt{x}\sqrt{1-\beta^2}\epsilon-\sqrt{1-x}\beta\right)^2}}\\ 
\beta &\defined 0\wedge \left(\min_{n\in[N]} \left<\ell_n, \ell\right> \,\,\text{s.t.}\,\, \left<\ell_n, \ell\right> > -1\right).
\]
\enlem
The proof of \cref{thm:gigabound} below follows by using the $\tau\sqrt{J_t}$ bound
from \cref{lem:geodesiclowerbound} to obtain an $O(1/t)$ bound on $J_t$, and then
combining that result with the $f(J_t)$ bound from \cref{lem:geodesiclowerbound} to obtain
the desired geometric decay.
\bprfof{\cref{thm:gigabound}}
Substituting the $\tau \sqrt{J_t}$ bound from \cref{lem:geodesiclowerbound} into \cref{eq:costrecursion} and
applying a standard inductive argument (e.g.~the proof of \citet[Lemma A.6]{Campbell17}) yields
\[
J_{t} \leq B(t)\defined \frac{J_1}{1+\tau^2J_1(t-1)}.\label{eq:Btbound}
\]
Since $B(t) \to 0$ and $f(B(t))\to\epsilon > 0$ as $t\to\infty$,
there exists a $t^\star\in\nats$ for which $t > t^\star$ implies $f(B(t)) \geq \tau \sqrt{B(t)}$.
Furthermore, since $f(x)$ is monotonically decreasing, we have that $f(J_t) \geq f(B(t))$.
Using the bound $\left<d_t, d_{tn_t}\right> \geq f(B(t))$ in \cref{eq:costrecursion} and combining with \cref{eq:Btbound} yields
\[
J_t \leq B(t\wedge t^\star)\prod_{s=t^\star+1}^{t} \left(1-f^2(B(s))\right).
\]
Multiplying by $\|\mcL\|^2$, taking the square root, and noting that $\eta$ from \cref{eq:etadefn} is equal to $\sqrt{J_1}$ gives the final result.
The convergence of $f(B(t)) \to \epsilon$ as $t\to\infty$ shows that the rate of decay
in the theorem statement is $\nu = \sqrt{1-\epsilon^2}$.
\eprfof

\section{Experiments}\label{sec:experiments}
In this section we evaluate the performance of GIGA coreset construction compared 
with both uniformly random subsampling and the Frank--Wolfe-based method of \citet{Campbell17}.
We first test the algorithms on simple synthetic examples, and then 
test them on logistic and Poisson regression models applied to numerous real and synthetic datasets.
Code for these experiments is available at 
\url{https://github.com/trevorcampbell/bayesian-coresets}.
\subsection{Synthetic Gaussian Inference}\label{sec:expt_gauss}
To generate \cref{fig:problem}, we generated  $\mu \dist \distNorm(0, 1)$
followed by $(y_n)_{n=1}^{10}\dist\distNorm(\mu, 1)$.
We then constructed Bayesian coresets via FW and GIGA using the norm specified in \citep[Section 6]{Campbell17}.
Across 1000 replications of this experiment, the median relative error in posterior variance approximation
is 3\% for GIGA and 48\% for FW.

\subsection{Synthetic Vector Sum Approximation}\label{sec:expt_synth}
In this experiment, we generated 20 independent datasets consisting of $10^6$ 50-dimensional vectors
from the multivariate normal distribution with mean 0 and identity covariance.
We then constructed coresets for each of the datasets via uniformly random subsampling (RND), Frank--Wolfe (FW), and GIGA.
We compared the algorithms on two metrics: reconstruction error, as measured by the 2-norm between $\mcL(w)$ and $\mcL$;
and representation efficiency, as measured by the size of the coreset. 

\begin{figure*}[t!]
\begin{center}
\begin{subfigure}[t]{0.47\textwidth}
\centering\includegraphics[width=.8\columnwidth]{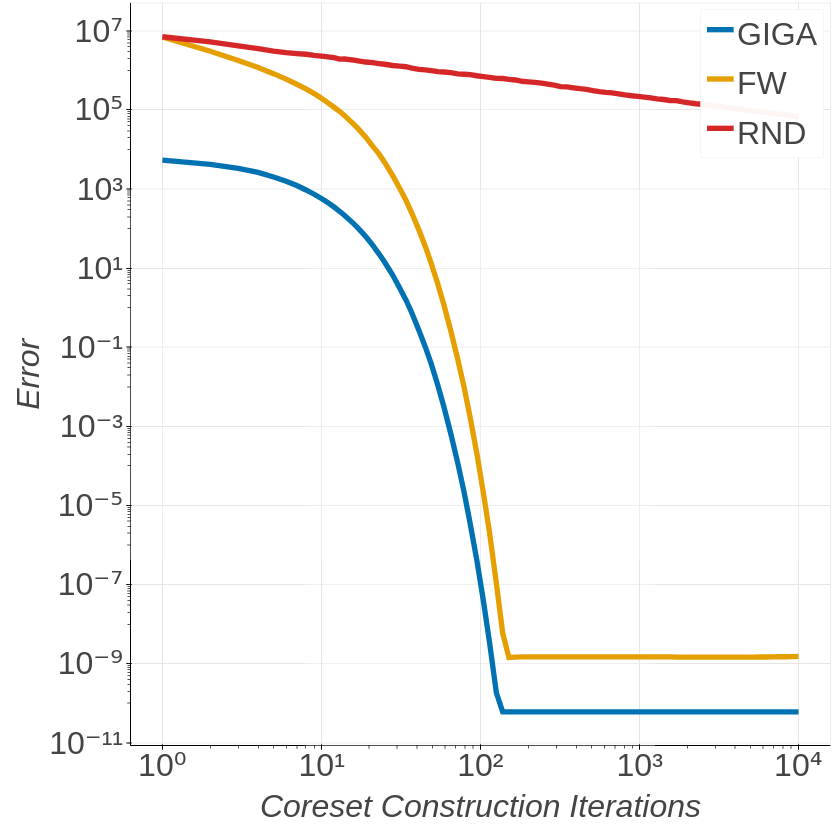}
\caption{}\label{fig:synth_err}
\end{subfigure}
\begin{subfigure}[t]{0.47\textwidth}
\centering\includegraphics[width=.8\columnwidth]{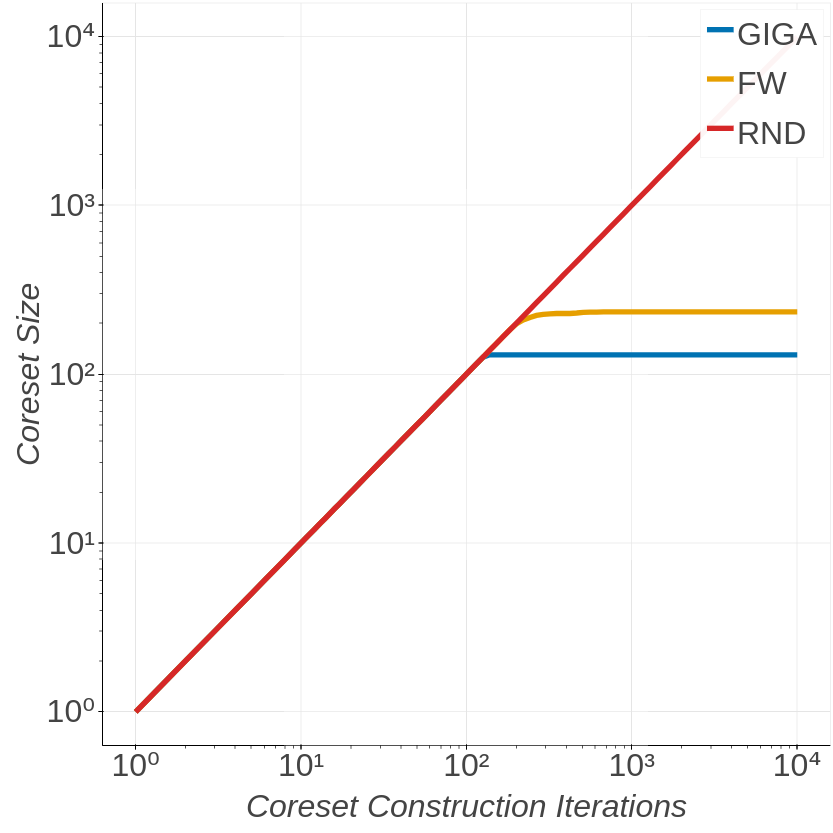}
\caption{}\label{fig:synth_csize}
\end{subfigure}
\end{center}
\caption{Comparison of different coreset constructions on the synthetic $\reals^{50}$ vector dataset.
\cref{fig:synth_err} shows a comparison of 2-norm error between the coreset $\mcL(w)$ and the true sum $\mcL$ as a function of construction iterations, demonstrating the significant improvement in quality when using GIGA.
\cref{fig:synth_csize} shows a similar comparison of coreset size, showing that GIGA produces smaller coresets for the same computational effort.
All plots show the median curve over 20 independent trials.}
\label{fig:synth}
\end{figure*}

\cref{fig:synth} shows the results of the experiment, with reconstruction error in \cref{fig:synth_err} and coreset size in \cref{fig:synth_csize}.
Across all construction iterations, GIGA provides a 2--4 order-of-magnitude reduction in error as compared with FW, and significantly outperforms RND. 
The exponential convergence of GIGA is evident. The flat section of FW/GIGA in \cref{fig:synth_err} for iterations beyond $10^2$ is due to the algorithm reaching the limits of numerical
precision. In addition, \cref{fig:synth_csize} shows that GIGA can improve representational efficiency over FW, ceasing to grow the coreset
once it reaches a size of 120, while FW continues to add points until it is over twice as large. Note that this experiment
is designed to highlight the strengths of GIGA: by the law of large numbers, as $N\to\infty$ 
the sum $\mcL$ of the $N$ \iid standard multivariate normal data vectors 
satisfies $\|\mcL\| \to 0$, while the sum of their norms $\sigma \to \infty$ \as
But \cref{sec:expt_ortho} shows that even in pathological cases, GIGA outperforms FW due to its optimal log-likelihood scaling.

\subsection{Bayesian Posterior Approximation}\label{sec:expt_bayes}
In this experiment, we used GIGA to generate Bayesian coresets for logistic and Poisson regression.
For both models, we used the standard multivariate normal prior $\theta \sim \distNorm(0, I)$.
In the \textbf{logistic regression} setting, we have a set of data points $\left(x_n, y_n\right)_{n=1}^N$ each consisting of a feature $x_n\in\reals^D$
and a label $y_n\in\{-1, 1\}$, and the goal is to predict the label of a new point  given its feature.
We thus seek to infer the posterior distribution of the parameter $\theta\in\reals^{D+1}$ governing the generation of $y_n$ given $x_n$
via
\[
\hspace{-.3cm}y_n \given x_n, \theta &\distind \distBern\left(\frac{1}{1+e^{-z_n^T\theta}}\right) \, z_n \defined \left[\!\!\begin{array}{c}x_n\\ 1\end{array}\!\!\right].
\]
In the \textbf{Poisson regression} setting, we have a set of data points $\left(x_n, y_n\right)_{n=1}^N$ each consisting of a feature $x_n\in\reals^D$
and a count $y_n\in\nats$, and the goal is to learn a relationship between features $x_n$ and the associated mean count.
We thus seek to infer the posterior distribution of the parameter $\theta\in\reals^{D+1}$ governing the generation of $y_n$ given $x_n$
via
\[
 \hspace{-.2cm}y_n \given x_n, \theta &\distind \distPoiss\left(\log\left(1+e^{z_n^T\theta}\right)\right) \, z_n \defined \left[\!\!\begin{array}{c}x_n\\ 1\end{array}\!\!\right].
\]

We tested the coreset construction methods for each model on a number of datasets; see \cref{sec:datasets} for references.
For \textbf{logistic regression}, the \texttt{Synthetic} dataset consisted of $N=$ 10,000 data points 
 with covariate $x_n \in \reals^2$ 
sampled \iid from $\distNorm(0, I)$, and label $y_n\in\{-1, 1\}$ generated from the logistic likelihood with parameter $\theta = \left[3, 3, 0\right]^T$.  
The \texttt{Phishing} dataset
consisted of $N=$ 11,055 data points 
each with $D=$ 68 features. 
The \texttt{DS1} dataset 
consisted of $N=$ 26,733 data points 
each with $D=$ 10 features. 

\begin{figure*}[t!]
\begin{center}
\begin{subfigure}[t]{0.47\textwidth}
\centering\includegraphics[width=.8\columnwidth]{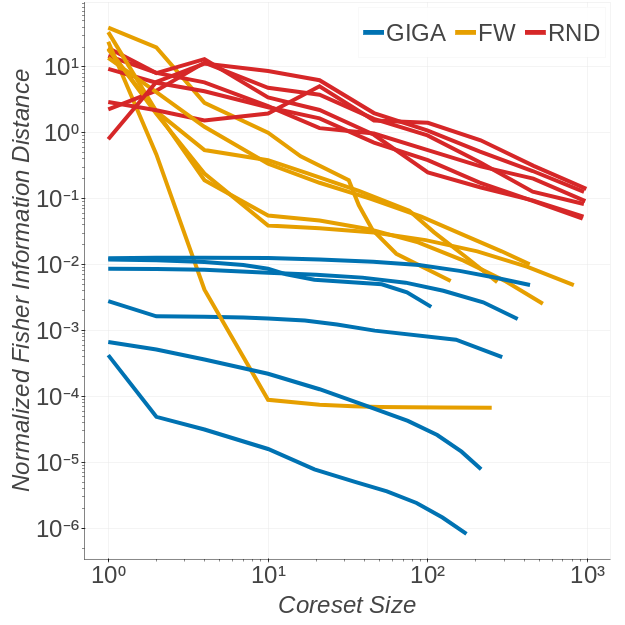}
\caption{}\label{fig:errvscsz}
\end{subfigure}
\begin{subfigure}[t]{0.47\textwidth}
\centering\includegraphics[width=.8\columnwidth]{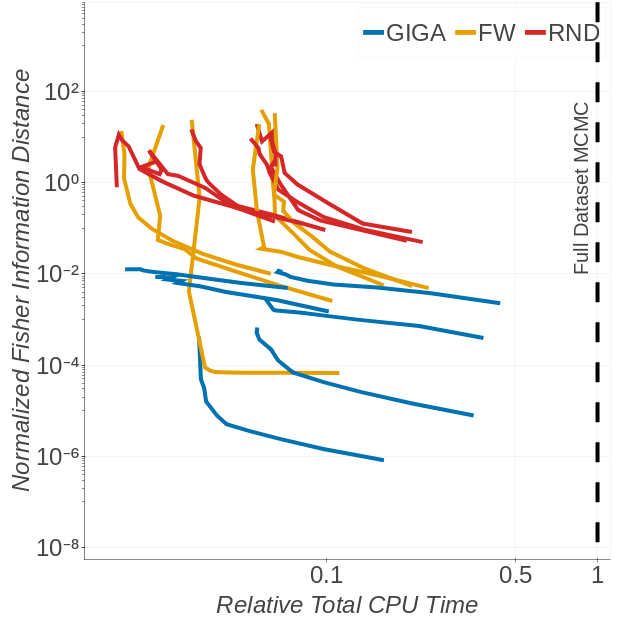}
\caption{}\label{fig:errvscput}
\end{subfigure}
\end{center}
\caption{Comparison of the median Fisher information distance to the true posterior for GIGA, FW, and RND on 
the logistic and Poisson regression models over 20 random trials. Distances are normalized by the median value 
of RND for comparison. In \cref{fig:errvscput}, computation time is normalized by the median value required to run 
MCMC on the full dataset. GIGA consistently outperforms FW and RND.}
\label{fig:lrpoiss}
\end{figure*}

For \textbf{Poisson regression}, the \texttt{Synthetic} dataset consisted of $N=$ 10,000 data points 
with covariate $x_n \in \reals$ 
sampled \iid from $\distNorm(0, 1)$, and count $y_n\in\nats$ generated from the Poisson likelihood with $\theta = \left[1, 0\right]^T$.
The \texttt{BikeTrips} dataset
consisted of $N=$ 17,386 data points 
each with $D=$ 8 features. 
The \texttt{AirportDelays} dataset 
consisted of $N=$ 7,580 data points 
each with $D=$ 15 features. 

We constructed coresets for each of the datasets via uniformly random subsampling (RND), Frank--Wolfe (FW), and GIGA 
using the weighted Fisher information distance,
\[
\|\mcL_n\|^2 \defined \EE_{\hat\pi}\left[\|\grad\mcL_n(\theta)\|_2^2\right],\label{eq:wFID}
\] 
where $\mcL_n(\theta)$ is the log-likelihood for data point $n$, and $\hat\pi$ is obtained via the Laplace approximation. 
In order to do so, we approximated all vectors $\mcL_n$ using a 500-dimensional random feature projection \citep[following][]{Campbell17}.
For posterior inference, we used Hamiltonian Monte Carlo \citep{Neal11} with 15 leapfrog steps per sample.
We simulated a total of 6,000 steps, with 1,000 warmup steps for step size adaptation with a target acceptance rate of 0.8, 
and 5,000 posterior sampling steps. \cref{sec:additional_expts} shows results for similar experiments using random-walk Metropolis--Hastings and the No-U-Turn Sampler \citep{Hoffman14}.
We ran 20 trials of projection / coreset construction / MCMC for each combination of dataset, model, and algorithm.
We evaluated the coresets at logarithmically-spaced construction iterations between $M=1$ and $1000$
by their median posterior Fisher information distance, estimated using samples obtained by running
posterior inference on the full dataset in each trial. We compared these results versus 
computation time as well as 
coreset size. The latter serves as an implementation-independent measure
of total cost, since the cost of running MCMC is much greater 
than running coreset construction and depends linearly on coreset size.

The results of this experiment are shown in \cref{fig:lrpoiss}. In \cref{fig:errvscsz}
and \cref{fig:errvscput}, the Fisher information distance is normalized by the median distance of RND
for comparison. In \cref{fig:errvscput}, the computation time is normalized by the median time to run MCMC
on the full dataset. The suboptimal coreset 
log-likelihood scaling of FW can be seen in \cref{fig:errvscsz} for small coresets,
resulting in similar performance to RND. In contrast, GIGA correctly scales posterior uncertainty across all coreset sizes, 
resulting in a major (3--4 orders of magnitude) reduction in error.
\cref{fig:errvscput} shows the same results plotted versus total computation time. 
This confirms that across a variety of models 
and datasets, GIGA provides significant improvements in posterior error over the state of the art.

\section{Conclusion}
This paper presented \emph{greedy iterative geodesic ascent (GIGA)}, a novel Bayesian coreset construction algorithm.
Like previous algorithms, GIGA is simple to implement, has low computational cost, and has no tuning parameters. But
in contrast to past work, GIGA scales the coreset log-likelihood optimally, providing 
significant improvements in the quality of posterior approximation.
The theoretical guarantees and experimental results presented in this work reflect this improvement. 

\section*{Acknowledgments}
This research is supported by an MIT Lincoln Laboratory Advanced
Concepts Committee Award, ONR grant N00014-17-1-2072, a Google
Faculty Research Award, and an ARO YIP Award.

{
\small
\bibliographystyle{icml2018}
\bibliography{main}
}

\newpage
\appendix
\begin{figure*}[t!]
\begin{center}
\begin{subfigure}[t]{0.47\textwidth}
\centering\includegraphics[width=.7\columnwidth]{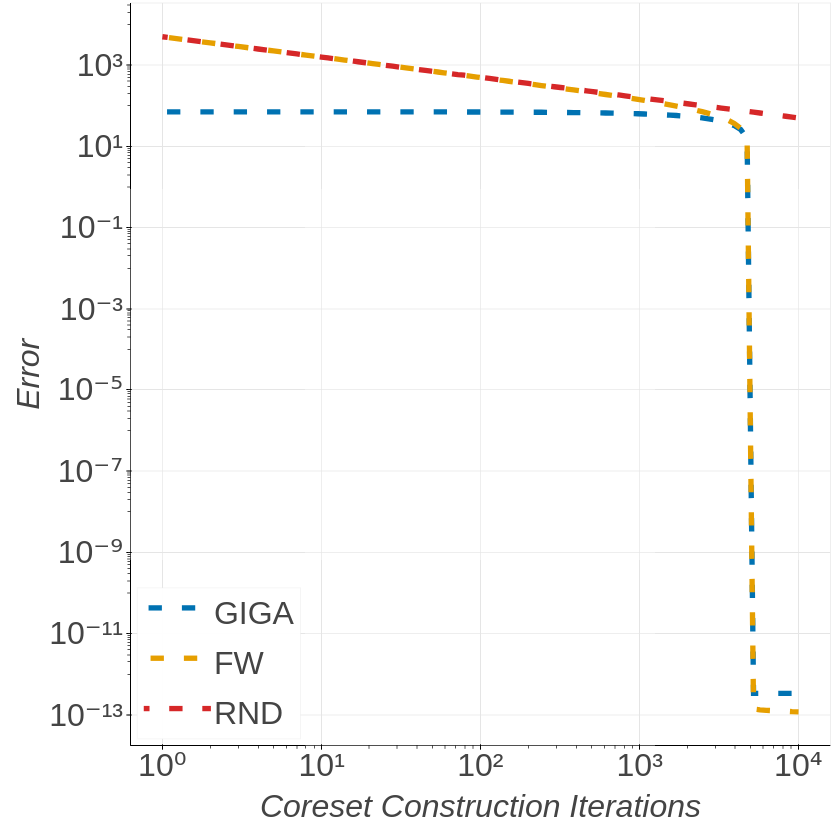}
\caption{}\label{fig:axis_err}
\end{subfigure}
\begin{subfigure}[t]{0.47\textwidth}
\centering\includegraphics[width=.7\columnwidth]{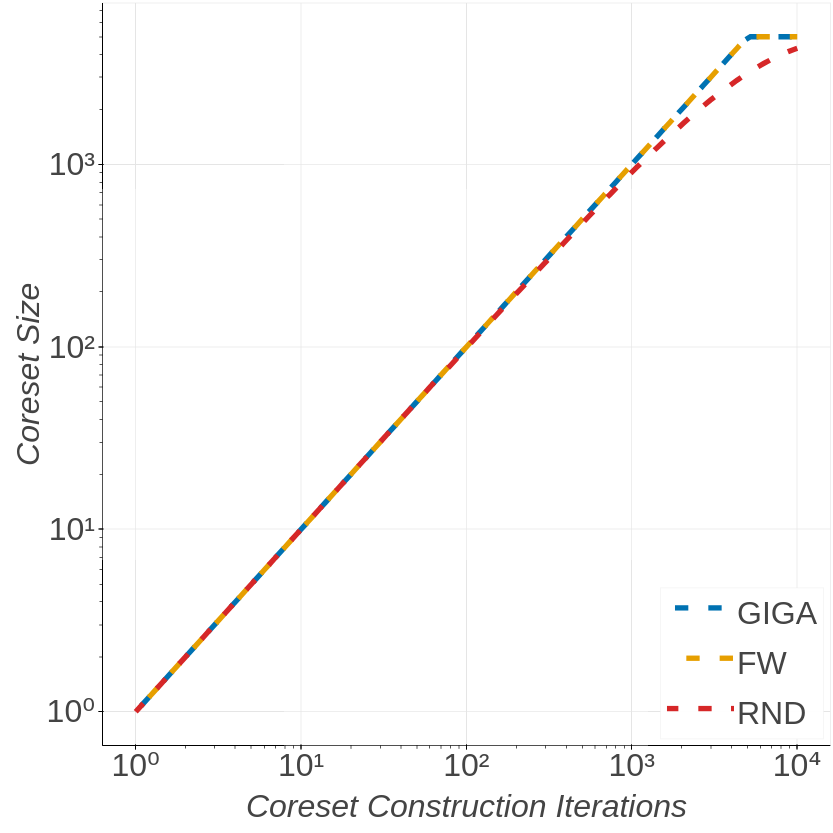}
\caption{}\label{fig:axis_csize}
\end{subfigure}
\end{center}
\caption{Comparison of different coreset constructions on the synthetic axis-aligned vector dataset.
\cref{fig:axis_err} shows a comparison of 2-norm error between the coreset $\mcL(w)$ and the true sum $\mcL$ as a function of construction iterations.
\cref{fig:axis_csize} shows a similar comparison of coreset size.}
\label{fig:axis}
\end{figure*}

\begin{figure*}[t!]
\begin{center}
\begin{subfigure}[t]{0.47\textwidth}
\centering\includegraphics[width=.7\columnwidth]{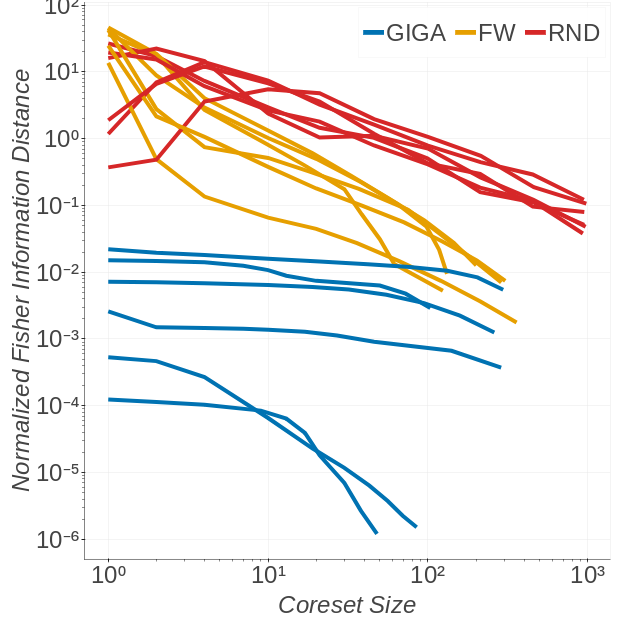}
\caption{}\label{fig:errvscsz_rwmh}
\end{subfigure}
\begin{subfigure}[t]{0.47\textwidth}
\centering\includegraphics[width=.7\columnwidth]{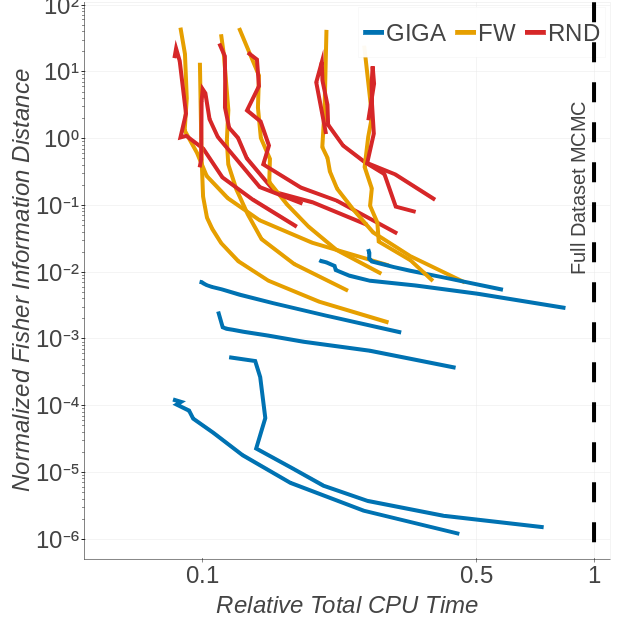}
\caption{}\label{fig:errvscput_rwmh}
\end{subfigure}
\end{center}
\caption{Results for the experiment described in \cref{sec:expt_bayes} with posterior inference via random-walk Metropolis--Hastings.}
\label{fig:lrpoiss_rwmh}
\end{figure*}

\begin{figure*}[t!]
\begin{center}
\begin{subfigure}[t]{0.47\textwidth}
\centering\includegraphics[width=.7\columnwidth]{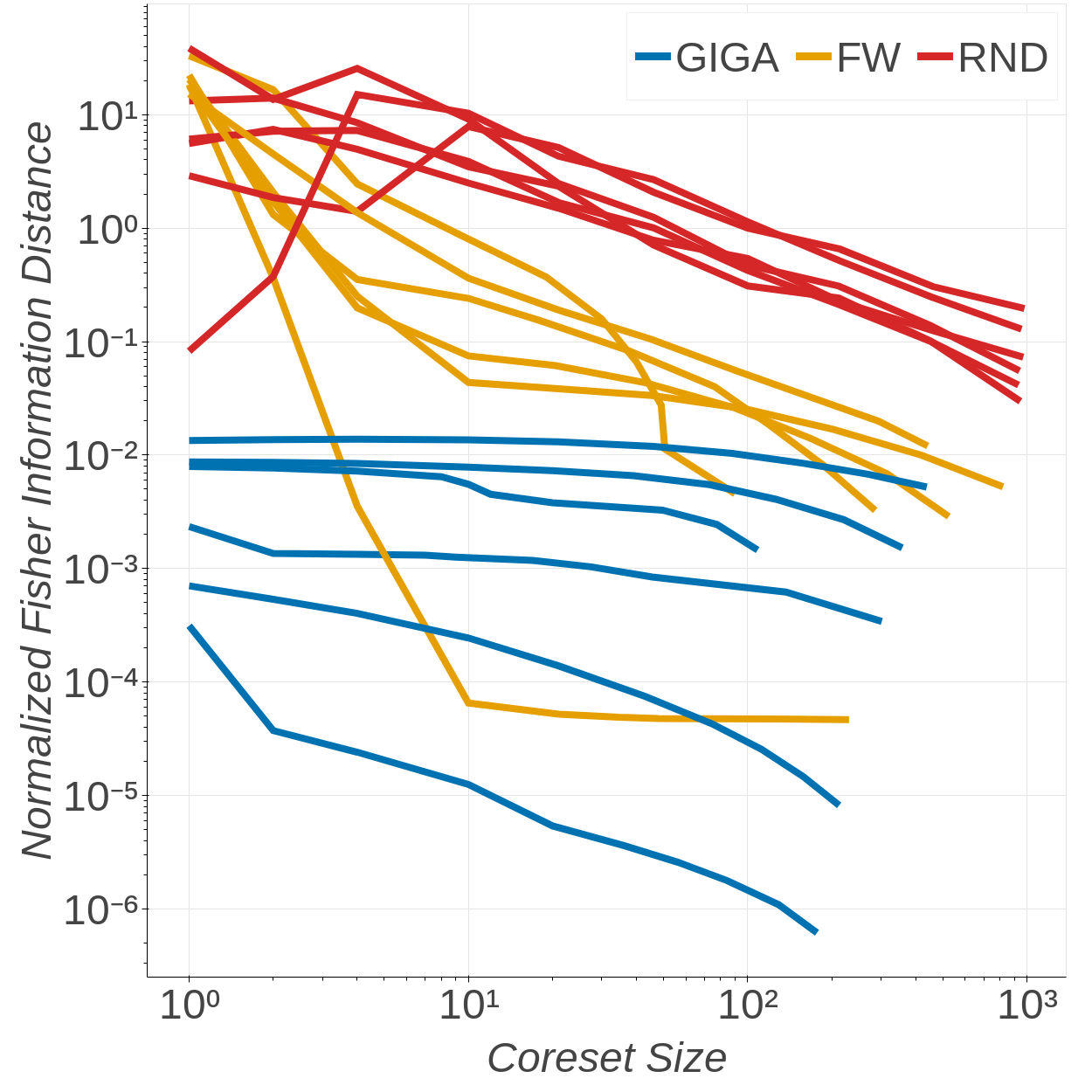}
\caption{}\label{fig:errvscsz_nuts}
\end{subfigure}
\begin{subfigure}[t]{0.47\textwidth}
\centering\includegraphics[width=.7\columnwidth]{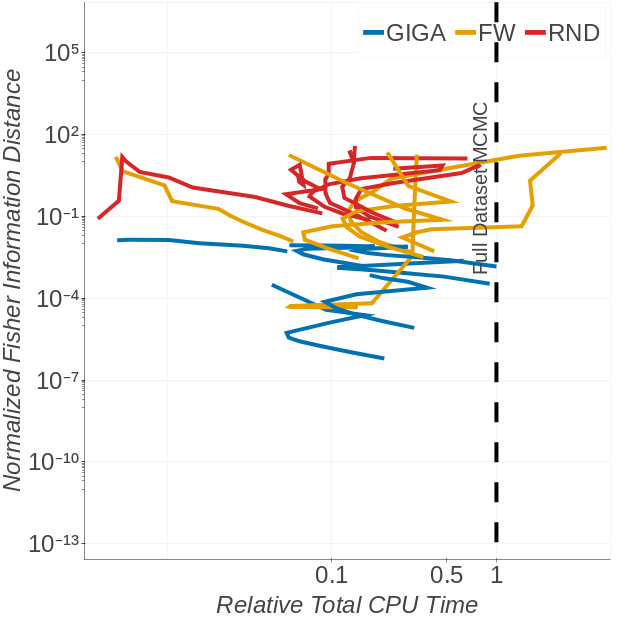}
\caption{}\label{fig:errvscput_nuts}
\end{subfigure}
\end{center}
\caption{Results for the experiment described in \cref{sec:expt_bayes} with posterior inference via NUTS.}
\label{fig:lrpoiss_nuts}
\end{figure*}

\section{Additional results}\label{sec:additional_expts}

\subsection{Orthonormal vectors}\label{sec:expt_ortho}

In this experiment, we generated a dataset of 5000 unit vectors in $\reals^{5000}$,
each aligned with one of the coordinate axes. This dataset is exactly that used
in the proof of \cref{prop:fwbad}, except that the number of datapoints $N$ is fixed to 5000.
We constructed coresets for each of the datasets via uniformly random subsampling (RND), Frank--Wolfe (FW), and GIGA.
We compared the algorithms on two metrics: reconstruction error, as measured by the 2-norm between $\mcL(w)$ and $\mcL$;
and representation efficiency, as measured by the size of the coreset. 
\cref{fig:axis} shows the results of the experiment, with reconstruction error in \cref{fig:axis_err} and coreset size in \cref{fig:axis_csize}.
As expected, for early iterations FW performs about as well as uniformly random subsampling, as these algorithms generate equivalent coresets 
(up to some reordering of the unit vectors) with high probability. FW only finds a good coreset after all 5000 points in the dataset have been added.
These algorithms both do not correctly scale the coreset; in contrast, GIGA scales its coreset correctly, providing significant reduction in error.

\subsection{Alternate inference algorithms}\label{sec:expt_other_inference}
We reran the same experiment as described in \cref{sec:expt_bayes},
except we swapped the inference algorithm for random-walk Metropolis--Hastings (RWMH)
and the No-U-Turn Sampler (NUTS) \citep{Hoffman14}.
When using RWMH, we simulated a total of 50,000 steps: 25,000 warmup steps
including covariance adaptation with a target acceptance rate of 0.234,
and 25,000 sampling steps thinned by a factor of 5, yielding 5,000 posterior samples.
If the acceptance rate for the latter 25,000 steps was not between 0.15 and 0.7, 
we reran the procedure.
When using NUTS, we simulated a total of 6,000 steps: 1,000 warmup steps
including leapfrog step size adaptation with a target acceptance rate of 0.8, 
and 5,000 sampling steps. 

The results for these experiments are shown in 
\cref{fig:lrpoiss_rwmh,fig:lrpoiss_nuts}, and generally corroborate the results
from the experiments using Hamiltonian Monte Carlo in the main text.
One difference when using NUTS is that the performance versus computation time
appears to follow an ``S''-shaped curve, which is caused by the dynamic path-length
adaptation provided by NUTS. Consider the log-likelihood of logistic regression,
which has a ``nearly linear'' region and a ``nearly flat'' region. When the coreset
is small, there are directions in latent space that point along ``nearly flat'' regions;
along these directions, u-turns happen only after long periods of travel.
When the coreset reaches a certain size, these ``nearly flat'' directions
are all removed, and u-turns happen more frequently. Thus we expect the computation time
as a function of coreset size to initially increase smoothly, then drop quickly, followed
by a final smooth increase, in agreement with  \cref{fig:errvscput_nuts}.

\section{Technical Results and Proofs}\label{sec:proofs}
\bprfof{\cref{lem:constantbounds}}
By setting $s_m = \frac{\|\mcL_m\|}{\|\mcL\|}$ for each $m\in[N]$ in \cref{eq:taudefn}, we have that $\tau \geq \frac{\|\mcL\|}{\sigma} > 0$.
Now suppose $\epsilon \leq 0$; then there exists some conic combination $d$ of $(d_{\infty m})_{m=1}^N$ for which $\|d\|=1$, $\left<d, \ell\right> = 0$, 
and $\forall m\in[N], \, \left<-d, d_{\infty m}\right> \leq 0$. There must exist at least one index $n\in[N]$ for which $\left<-d, d_{\infty n}\right> < 0$, since 
otherwise $d$ is not in the linear span of  $(d_{\infty m})_{m=1}^N$. This also implies $\|d_{\infty n}\| > 0$ and hence $\|\ell_n - \left<\ell_n, \ell\right>\ell\| > 0$.
Then $\left<-d, \sum_{m=1}^N \xi_m d_{\infty m}\right> < 0$ for any $\xi\in\Delta^{N-1}$ with $\xi_n > 0$.
But setting $\xi_n \propto \sigma_n \|\ell_n - \left<\ell_n, \ell\right>\ell\|$ results in $\sum_{n=1}^N \xi_n d_{\infty n} = 0$, and we have a contradiction.
\eprfof

\bprfof{\cref{lem:geodesiclowerbound}}
We begin with the $\tau\sqrt{J_t}$ bound. For any $\xi\in\Delta^{N-1}$,
\[
\left<d_t, d_{tn_t}\right> &= \max_{n\in[N]}\left<d_t, d_{tn}\right> \geq \sum_{n=1}^N\xi_n\left<d_t, d_{tn}\right>.
\]
Suppose that $\ell = \sum_{n=1}^N s_n\ell_n$ for some $s \in \reals^N_+$.
Setting $\xi_n \propto s_n\left\|\ell_n - \left<\ell_n, \ell(w_t)\right>\ell(w_t)\right\|$ yields
\[
\left<d_t, d_{tn_t}\right> &\geq C^{-1}\left\|\ell - \left<\ell, \ell(w_t)\right>\ell(w_t)\right\|\\
C &\defined \left(\sum_{n=1}^Ns_n\left\|\ell_n - \left<\ell_n, \ell(w_t)\right>\ell(w_t)\right\|\right).
\]
Noting that the norms satisfy $\left\|\ell - \left<\ell, \ell(w_t)\right>\ell(w_t)\right\| = \sqrt{J_t}$
and $\left\|\ell_n - \left<\ell_n, \ell(w_t)\right>\ell(w_t)\right\| \leq 1$, we have
\[
\left<d_t, d_{tn_t}\right> &\geq \|s\|_1^{-1}\sqrt{J_t}.
\]
Maximizing over all valid choices of $s$ yields
\[
\left<d_t, d_{tn_t}\right> &\geq \tau\sqrt{J_t}.
\]
Next, we develop the $f(J_t)$ bound. Note that 
\[
\sum_{n=1}^N w_{tn} \left\|\ell_n - \left<\ell_n, \ell\right>\ell\right\| d_{\infty n}
 &= \sum_{n=1}^Nw_{tn} (\ell_n - \left<\ell_n, \ell\right>\ell)\\
&= \ell(w_t) - \left<\ell(w_t), \ell\right>\ell,
\]
so  we can express
$\ell(w_t) = \sqrt{J_t} d + \sqrt{1-J_t}\ell$ and $d_t = \sqrt{J_t} \ell - \sqrt{1-J_t} d$ for some vector $d$ that 
is a conic combination of $\left(d_{\infty n}\right)_{n=1}^N$ with $\|d\|=1$ and $\left<d, \ell\right> = 0$.
Then by the definition of $\epsilon$ in \cref{eq:epsdefn} and \cref{lem:constantbounds}, there exists an $n\in[N]$ such that $\left<-d, d_{\infty n}\right> \geq \epsilon > 0$.
Therefore
{\small
\[
&\left<d_t, d_{tn_t}\right> \\
&\geq \left<d_t, d_{tn}\right>\\
&= \left<\sqrt{J_t}\ell-\sqrt{1-J_t}d, \frac{\ell_n - \left<\ell_n, \ell(w_t)\right>\ell(w_t)}{\|\ell_n - \left<\ell_n, \ell(w_t)\right>\ell(w_t)\|}\right>\\
&= \frac{\sqrt{1-J_t}\left<-d, \ell_n\right> + \sqrt{J_t}\left<\ell, \ell_n\right>}{\sqrt{1-\left(\sqrt{1-J_t}\left<\ell_n, \ell\right>+\sqrt{J_t}\left<\ell_n, d\right> \right)^2}}\\
&=\frac{\sqrt{1-J_t}\sqrt{1-\left<\ell_n, \ell\right>^2} \left<-d, d_{\infty n}\right> + \sqrt{J_t}\left<\ell, \ell_n\right>}{\sqrt{1-\left(\sqrt{1-J_t}\left<\ell_n, \ell\right>+\sqrt{J_t}\sqrt{1-\left<\ell_n,\ell\right>^2}\left<d,d_{\infty n}\right>\right)^2}}.
\]}
We view this bound as a function of two variables $\left<\ell, \ell_n\right>$ and $\left<-d, d_{\infty n}\right>$, and we view the worst-case bound as the minimization over these variables.
We further lower-bound by removing the coupling between them. 
Fixing $\left<-d, d_{\infty n}\right>$, the derivative in $\left<\ell, \ell_n\right>$ is always nonnegative, 
and note that $\left<\ell_n, \ell\right> > -1$ since otherwise $\left<-d, d_{\infty n}\right> = 0$ by the remark after \cref{eq:selection},
so setting 
\[
\beta = 0 \wedge \left(\min_{n\in[N]}\left<\ell, \ell_n\right> \,\,\text{s.t.}\,\,\left<\ell, \ell_n\right> > -1\right),
\] 
we have
{\small
\[
&\left<d_t, d_{tn_t}\right> \geq \\
&\frac{\sqrt{1-J_t}\sqrt{1-\beta^2} \left<-d, d_{\infty n}\right> + \sqrt{J_t}\beta}{\sqrt{1-\left(\sqrt{1-J_t}\beta+\sqrt{J_t}\sqrt{1-\beta^2}\left<d,d_{\infty n}\right>\right)^2}}.
\]}
We add $\{0\}$ into the minimization since $\beta\leq 0$ guarantees that the derivative of the above with respect to $J_t$ is nonpositive (which we will require in proving the main theorem).
For all $J_t$ small enough such that $\sqrt{1-J_t}\sqrt{1-\beta^2} \epsilon + \sqrt{J_t}\beta \geq 0$, the derivative of the above with respect to $\left<-d, d_{\infty n}\right>$
is nonnegative. Therefore, minimizing yields
\[
\left<d_t, d_{tn_t}\right> \geq 
\frac{\sqrt{1-J_t}\sqrt{1-\beta^2} \epsilon + \sqrt{J_t}\beta}{\sqrt{1-\left(\sqrt{1-J_t}\beta-\sqrt{J_t}\sqrt{1-\beta^2}\epsilon\right)^2}}.
\]
which holds for any such small enough $J_t$. But note that we've already proven the $\left<d_t, d_{tn_t}\right>\geq \tau\sqrt{J_t}$ bound, which is always nonnegative; so the only time 
the current bound is ``active'' is when it is itself nonnegative, i.e.~when $J_t$ is small enough. Therefore the bound
\[
\textstyle \left<d_t, d_{tn_t}\right> \geq 
\tau\sqrt{J_t}\vee 
\frac{\sqrt{1-J_t}\sqrt{1-\beta^2} \epsilon + \sqrt{J_t}\beta}{\sqrt{1-\left(\sqrt{1-J_t}\beta-\sqrt{J_t}\sqrt{1-\beta^2}\epsilon\right)^2}}
\]
holds for all $J_t \in [0, 1]$.
\eprfof

\section{Cap-tree Search}\label{sec:balltree}
When choosing the next point to add to the coreset, we need to solve the following maximization with $O(N)$ complexity:
\[
n_t &= \argmax_{n\in[N]} \frac{\left<\ell_n, \ell-\left<\ell, \ell(w_t)\right>\ell(w_t)\right>}{\sqrt{1-\left<\ell_n, \ell(w_t)\right>^2 }}.
\]
One option to potentially reduce this complexity is to first partition the data in a tree structure, and use the tree structure for faster search.
However, we found that in practice (1) the cost of constructing the tree structure outlined below outweighs the benefit of faster search later on, and (2)
the computational gains diminish significantly with high-dimensional vectors $\ell_n$. 
We include the details of our proposed cap-tree below, and leave more efficient construction and search as an open problem for future work.

Each node in the tree is a spherical ``cap'' on the surface of the unit sphere, defined by a central direction $\xi$, $\|\xi\|=1$ and a dot-product bound $r\in[-1, 1]$, with the property that 
all data in child leaves of that node satisfy $\left<\ell_n, \xi\right> \geq r$. Then we can upper/lower bound the search objective for such data given $\xi$ and $r$.
If we progress down the tree, keeping track of the best lower bound, we may be able to prune large quantities of data if the upper bound of any node is less than the current best lower bound.

For the lower bound, we 
evaluate the objective at the vector $\ell_n$ closest to $\xi$. 
For the upper bound, define $u \defined \frac{\ell-\left<\ell, \ell(w_t)\right>\ell(w_t)}{\|\ell-\left<\ell, \ell(w_t)\right>\ell(w_t)\|}$,
and $v \defined \ell(w_t)$. Then $\|u\|=\|v\|=1$ and $\left<u, v\right> = 0$.
The upper bound is
\[
\max_{\zeta} \frac{\left<\zeta, u\right>}{\sqrt{1-\left<\zeta, v\right>^2 }}  \quad
\text{s.t.}  \quad  \left<\zeta, \xi\right> \geq r \quad
                    \|\zeta\|= 1.
\]
If we write $\zeta = \alpha_u u + \alpha_v v + \sum_i \alpha_i z_i$
where $z_i$ completes the basis of $u, v$ etc, and
$\xi = \beta_u u + \beta_v v + \sum_i \beta_i z_i$,
\[
\max_{\alpha\in\reals^d}\quad & \frac{\alpha_u}{\sqrt{1-\alpha_v^2 }}\\ 
\text{s.t.} \quad &    \alpha_u\beta_u + \alpha_v\beta_v + \sum_i \alpha_i\beta_i \geq r\notag\\
 & \alpha_u^2+\alpha_v^2+\sum_i \alpha_i^2 = 1.\notag
\]
Noting that $\alpha_i$ doesn't appear in the objective, we maximize $\sum_i \alpha_i\beta_i$ to find the equivalent optimization
\[
\max_{\alpha_u, \alpha_v} \quad& \frac{\alpha_u}{\sqrt{1-\alpha_v^2 }} \\ 
\text{s.t.}  \quad&  \alpha_u\beta_u + \alpha_v|\beta_v| + \|\beta\|\sqrt{1-\alpha_u^2-\alpha_v^2} \geq r\\
 &\alpha_u^2+\alpha_v^2 \leq 1,
\]
where the norm on $|\beta_v|$ comes from the fact that we can choose the sign of $\alpha_v$ arbitrarily, ensuring the optimum has $\alpha_v \geq 0$.
Now define 
\[
\gamma \defined \frac{\alpha_u}{\sqrt{1-\alpha_v^2}} \quad \eta \defined \frac{1}{\sqrt{1-\alpha_v^2}},
\]
so that the optimization becomes
\[
\max_{\gamma, \eta} \quad& \gamma\\
\text{s.t.}  \quad&  \gamma\beta_u + |\beta_v|\sqrt{\eta^2-1} + \|\beta\|\sqrt{1-\gamma^2} \geq r\eta\\
 &\gamma^2 \leq 1, \, \eta \geq 1.
\]
Since $\eta$ is now decoupled from the optimization, we can solve
\[
\max_{\eta \geq 1} |\beta_v|\sqrt{\eta^2-1} - r\eta \label{eq:etaopt}
\]
to make the feasible region in $\gamma$ as large as possible.
If $|\beta_v| > r$, we maximize \cref{eq:etaopt} by sending $\eta \to \infty$ yielding a maximum of 1 in the original optimization.
Otherwise, note that at $\eta = 1$ the derivative of the objective is $+\infty$, so we know the constraint $\eta = 1$ is not active.
Therefore, taking the derivative and setting it to 0 yields
\[
0&=\frac{|\beta_v|\eta}{\sqrt{\eta^2-1}}-r\\
\eta &= \sqrt{\frac{r^2}{r^2-|\beta_v|^2}}.
\]
Substituting back into the original optimization,
\[
\max_{\gamma} \quad& \gamma  \\
\text{s.t.}  \quad&  \gamma\beta_u + \|\beta\|\sqrt{1-\gamma^2} \geq \sqrt{r^2-|\beta_v|^2}\\
& \gamma^2 \leq 1.
\]
If $\beta_u \geq \sqrt{r^2-|\beta_v|^2}$, then $\gamma = 1$ is feasible and the optimum is 1.
Otherwise,
note that at $\gamma = -1$, the derivative of the constraint is $+\infty$ and the derivative of the objective is $1$,
so the constraint $\gamma = -1$ is not active. Therefore, we can solve the unconstrained optimization
by taking the derivative and setting to 0, yielding
\[
\gamma = \frac{\beta_u\sqrt{r^2-\beta_v^2}+\|\beta\|\sqrt{1-r^2}}{\|\beta\|^2+\beta_u^2}.
\]
Therefore, the upper bound is as follows:
\[
U = \left\{\begin{array}{ll}
   1 & |\beta_v| > r\\
   1 & \beta_u \geq \sqrt{r^2-\beta_v^2}\\
  \frac{\beta_u\sqrt{r^2-\beta_v^2}+\|\beta\|\sqrt{1-r^2}}{\|\beta\|^2+\beta_u^2} & \text{else.}
  \end{array}\right.
\]

\section{Datasets}\label{sec:datasets}
The \texttt{Phishing} dataset is available online at \url{https://www.csie.ntu.edu.tw/~cjlin/libsvmtools/datasets/binary.html}.
The \texttt{DS1} dataset is available online at  \url{http://komarix.org/ac/ds/}.
The \texttt{BikeTrips} dataset is available online at \url{http://archive.ics.uci.edu/ml/datasets/Bike+Sharing+Dataset}.
The \texttt{AirportDelays} dataset was constructed using flight delay data from \url{http://stat-computing.org/dataexpo/2009/the-data.html}
and historical weather information from \url{https://www.wunderground.com/history/}.

\end{document}